\documentclass[a4paper]{article}

\usepackage{INTERSPEECH_v2}
\usepackage{dsfont}
\usepackage{graphicx}
\usepackage{xcolor}
\usepackage{caption}
\usepackage{subcaption}
\usepackage{multirow}
\usepackage{color,soul}

\title{Progressive Neural Networks for Transfer Learning in Emotion Recognition}
  
\name{John Gideon$^{*1}$, Soheil Khorram$^{*1}$, Zakaria Aldeneh$^{1}$, 
\\Dimitrios Dimitriadis$^2$, Emily Mower Provost$^{1}$}
\address{$^1$University of Michigan at Ann Arbor, $^2$IBM T. J.  Watson Research Center}
\email{\texttt{\{gideonjn, khorrams, aldeneh, emilykmp\}@umich.edu,
        dbdimitr@us.ibm.com}}
\begin{document}

\maketitle


\begin{abstract}
  Many paralinguistic tasks are closely related and thus representations learned 
  in one domain can be leveraged for another. In this paper, we 
  investigate how knowledge can be transferred between three paralinguistic 
  tasks: speaker, emotion, and gender recognition. Further,
  we extend this problem to cross-dataset tasks, asking how knowledge captured in one emotion dataset 
  can be transferred
  to another. We focus on progressive neural networks and compare these networks to the conventional deep learning
  method of pre-training 
  and fine-tuning. Progressive neural networks provide a way to transfer 
  knowledge and avoid the forgetting effect present when 
  pre-training neural networks on different tasks. Our experiments demonstrate 
  that: (1) emotion recognition can benefit from using representations originally learned 
  for different paralinguistic tasks and (2) transfer learning can
  effectively leverage additional datasets to improve the performance of emotion 
  recognition systems.
\end{abstract}
\noindent\textbf{Index Terms}: neural networks, transfer learning, 
progressive neural networks, computational paralinguistics, emotion recognition
  
\makeatletter{\renewcommand*{\@makefnmark}{}
	\footnotetext{$^*$These authors contributed equally to this work}}


\section{Introduction}
  Automatic emotion recognition has been actively explored by 
  researchers for the past few decades. However, compared with other 
  tasks such as automatic speech recognition (ASR) or speaker verification, 
  the sizes of emotion 
  datasets are relatively small. This makes it difficult to create models 
  that generalize beyond the recording conditions and subject demographics of a particular dataset.
  One possible approach to alleviate this problem is to incorporate related knowledge that can help in learning
  a better system.
  Many paralinguistic tasks are closely related and dependent on one another. 
  As a result, we hypothesize that emotion recognition models can be augmented with additional information
  from other paralinguistic information, such as speaker 
  ID and gender, to improve classification performance. 
  In this paper, 
  we explore transfer learning both as a method to leverage knowledge from other 
  paralinguistic tasks and to augment an emotion model with another model trained 
  on a different emotion dataset.  
  
    
  Previous work demonstrated that multi-task learning techniques (MTL) could be used to jointly model
  both gender and emotion~\cite{ververidis2004automatic,lee2005toward,vogt2006improving}, resulting in consistent performance increases compared to
  gender-agnostic models. Previous work also showed that emotion recognition systems could be improved by incorporating speaker identity as a feature, along with the other emotion-related features~\cite{sidorov2014comparison}. It has also been demonstrated that some features (e.g., pitch) are not only valuable for predicting emotions, but are also effective for detecting weight, height, gender, etc.~\cite{schuller2013paralinguistics}.  
Zhang et al.~\cite{zhang2016cross} explored MTL frameworks for leveraging data from different domains (speech and song) and gender in emotion recognition systems. Xia et al.~\cite{xia2015multi} treated dimensional and categorical labels as two different tasks.

MTL is best suited to situations where it is possible to train with all data from scratch. However, in some cases an existing model needs to be adapted to a new situation.
  Transfer learning provides a framework from which to address this problem.
  The most common approach to transfer learning is to train a model in one domain and fine-tune it in a related domain \cite{das2015cross, deng2013sparse, ng2015deep}. This pre-training and fine-tuning (PT/FT) approach has been successful in cases where the size of available data for the source domain is abundant in comparison to size of the data in the target domain. For example, in ASR, transfer learning is used to transfer knowledge from a richly-resourced language to an under-resourced language \cite{das2015cross}. In emotion recognition, Deng et al.~\cite{deng2013sparse} presented a sparse auto-encoder method for transferring knowledge between six emotion recognition datasets. The authors used auto-encoders to transform a source domain and its features to a domain that is more consistent with the target domain. The authors then used the transformed source domain to learn SVM classifiers to predict emotions in the target domain. Ng et al.~\cite{ng2015deep} studied transfer learning in the context of facial expression recognition. The authors first trained a model on a general large-scale dataset (ImageNet) that contains $1.2$ million images. They then fine-tuned the trained model on four datasets that contain relevant facial expressions similar to those in the target domain. Finally, the authors fine-tuned the network using the target data. The authors found that the two-step fine tuning approach provided improvements over one-step fine-tuning and no fine-tuning.
  
 However, the PT/FT approach has a number of limitations. First, it is unclear how 
 to initialize a model given learned weights from a sequence of related  tasks
 \cite{rusu2016progressive}. Second, when a model is 
 fine-tuned using initial weights learned from a source task, the end-model loses
its ability to solve the source task, a phenomenon termed the ``forgetting effect''~\cite{fahlman1990cascade}.
 Finally, transferring learned parameters
 between networks can be challenging if the networks have inconsistent architectures.
 In this work, we use the PT/FT approach as our baseline for
 transfer learning.

  Another recent approach for transfer learning is progressive neural networks (ProgNets) \cite{rusu2016progressive, rusu2016sim}. ProgNets train sequences of tasks by freezing the previously trained tasks and using their intermediate representations as inputs into the new network. This allows ProgNets to overcome the above-mentioned limitations associated with the traditional method of PT/FT, including the challenge of initializing a model from a sequence of models, at the expense of added parameters. Additionally, it prevents the forgetting effect present in the PT/FT methods by freezing and preserving the source task weights.
  
  In this work, we investigate transfer learning between three
  paralinguistic tasks: emotion, speaker, and gender 
  recognition, with a focus on emotion recognition as the target domain. 
  In addition, we investigate the efficacy of transfer learning applied between two
  emotion datasets: IEMOCAP~\cite{busso2008iemocap} and MSP-IMPROV~\cite{busso2016msp}. 
  Finally, we study the effect of transfer learning between datasets when the target task
  has limited amount of data available.
  In all cases, we investigate three methods: (1) deep neural network (DNN); (2) DNN 
  with PT/FT; and (3) progressive networks. Our results demonstrate significant improvements over the conventional PT/FT methods when using ProgNets for transferring knowledge from speaker recognition to emotion recognition tasks. Furthermore, our results suggest that ProgNets show promise as a method for transferring knowledge from gender detection to emotion recognition tasks, as well as transferring knowledge across datasets, with results significantly better than a DNN without transfer learning.
  

\section{Datasets and Features}
\subsection{Datasets}
We use speech utterances from two datasets in our study: 
IEMOCAP~\cite{busso2008iemocap} and MSP-IMPROV~\cite{busso2016msp}, two of the most commonly used datasets in emotion classification \cite{kim2016emotion,lotfian2016retrieving}. Both datasets were collected to simulate natural dyadic interactions
between actors and have similar labeling schemes. We use utterances with majority agreement
ground-truth labels. We only consider utterances with happy, sad, angry, 
and neutral labels.

\textbf{IEMOCAP:} The IEMOCAP dataset contains utterances from ten speakers 
grouped into five sessions. Each session contains one male and one female 
speaker. We combine excitement 
and happiness utterances to form the happy category, as in~\cite{busso2008iemocap}. The final dataset
contains 5531 utterances (1103 angry, 1708 neutral, 1084 sad, 1636 happy).

\textbf{MSP-IMPROV:} The MSP-IMPROV dataset contains utterances from 12 
speakers grouped into six sessions. A session has one male and one 
female. The final dataset contains 7798 utterances (792 angry, 
3477 neutral, 885 sad, 2644 happy).

\subsection{Features}
We use the eGeMAPS~\cite{eyben2016geneva} feature set
designed to standardize features used in affective computing. The eGeMAPS feature 
set contains a total of $88$ features, including frequency, energy, spectral, cepstral, and dynamic information. The final feature vectors for each
utterance are obtained by applying the following statistics: mean, coefficient
of variation, 20-th, 50-th, and 80-th percentile, range of 20-th to 80-th 
percentile, mean and standard deviation of the slope of rising/falling signal 
parts, mean of the Alpha Ratio, the 
Hammarberg Index, and the spectral slopes from $0$--$500$Hz and $500$--$1500$Hz. 
We perform dataset-specific global $z$-normalization on all features.
 

\begin{figure}[t]
\centering \includegraphics[width=\linewidth]{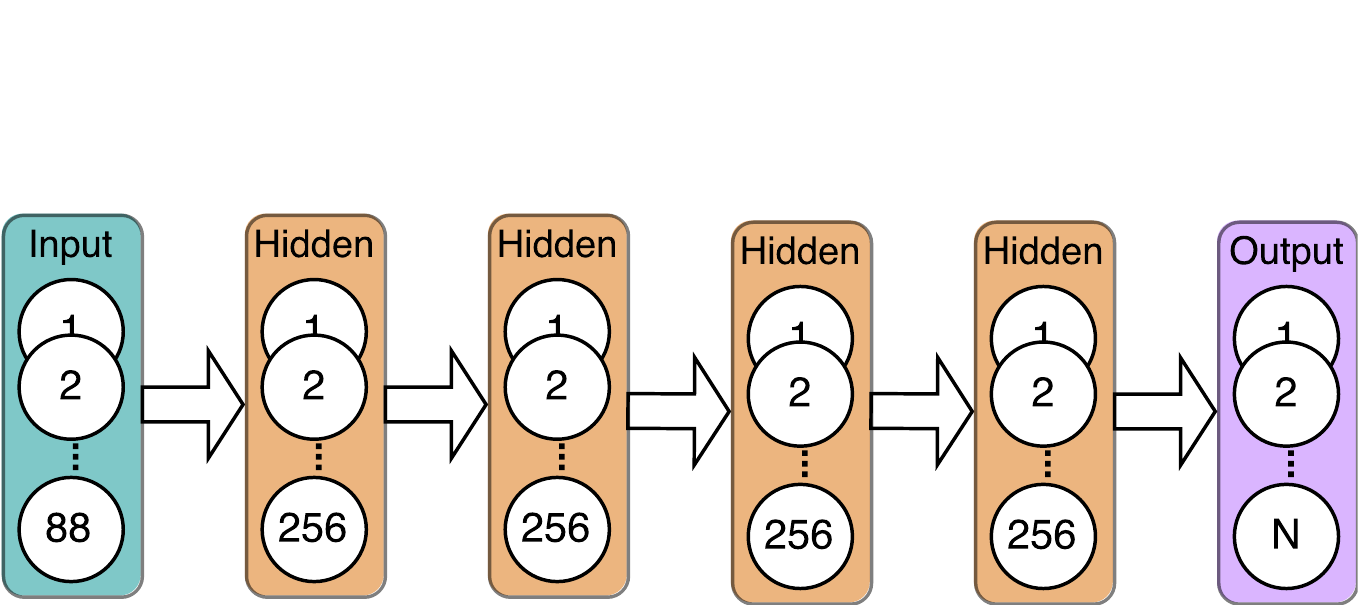}
\caption{Deep Neural Network (DNN) used in our experiments. The arrows represent 
dense connections between each layer. The number of outputs (N) varies depending 
on the experiment.}
\label{dnnFigure}
\end{figure}

\begin{figure}[t]
\centering \includegraphics[width=\linewidth]{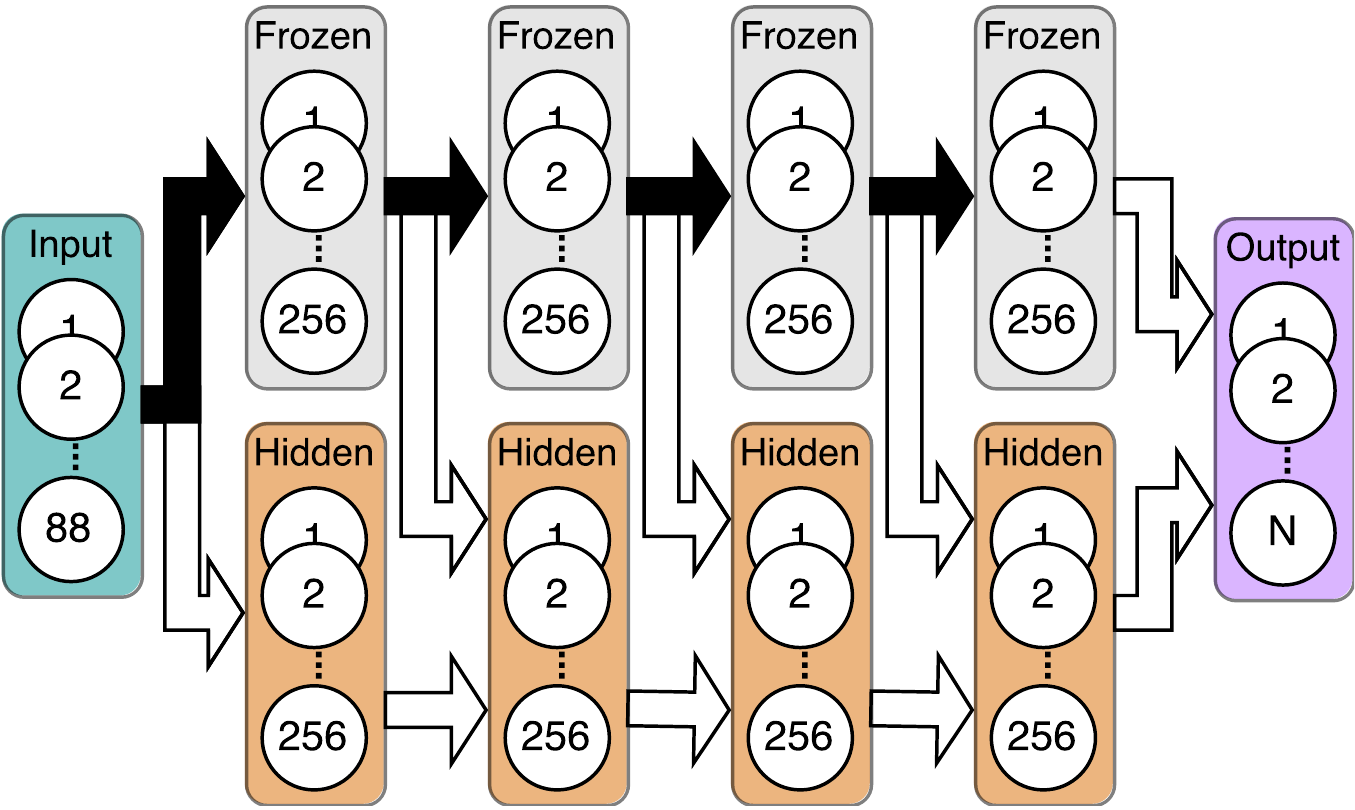}
\caption{Progressive Neural Network (ProgNet) used in our experiments. The arrows
represent dense connections between each layer. The black arrows show frozen 
weights from the transferred representations. The number of outputs (N) varies 
depending on the experiment.}
\label{progNetFigure}
\end{figure}

\section{Methods}
 We compare three methods in the context of transfer learning. As a baseline method, we consider the performance of a DNN trained on the target task without any extra knowledge. Additionally, we use the common transfer learning approach of pre-training a DNN on the source task and fine-tuning on the target task (PT/FT).
 The underlying assumption of PT/FT is that the target model can leverage
 prior knowledge present in the source task. This approach has been effective in many
 applications, including ASR~\cite{das2015cross} and 
 natural language processing~\cite{mou2016transferable}.

Both these methods are compared to the recently introduced progressive neural networks (ProgNets) \cite{rusu2016progressive}. 
Instead of using learned parameters as a starting point for
training a model on a target task,
ProgNets do the following: (1) freeze all parameters of the old model; (2) add
a new model that is initialized randomly; 
(3) add connections between the old (frozen) model and the new model; (4) learn 
parameters of the new model using backpropagation. ProgNets do not disrupt the learned information in existing source tasks, which avoids the forgetting effect present in PT/FT \cite{rusu2016progressive}.

In the construction of ProgNets, it is important to carefully select a method for combining representations across network and to identify where these representations will be combined. Adaptation layers can be included to transform from one task's representation to another.
However, due to the small amount of data available for training, we use ProgNets without adaptation layers. For the same reason, we simplify the network by using an equal number of layers in each column and transfer the representations between neighboring layers in a one-to-one fashion: the representations produced at layer $k$ from the
frozen column is fed as an input to layer $k+1$ of the new column
(Figure~\ref{progNetFigure}).

Table \ref{params_table} shows the neural network 
parameters used by all experiments in this paper. These values were selected using a standard DNN without transfer learning to determine the best structure suited to our data (Figure~\ref{dnnFigure}).
\begin{figure*}[t]
\centering
\begin{subfigure}[b]{0.46\linewidth}
\centering \includegraphics[width=\linewidth]{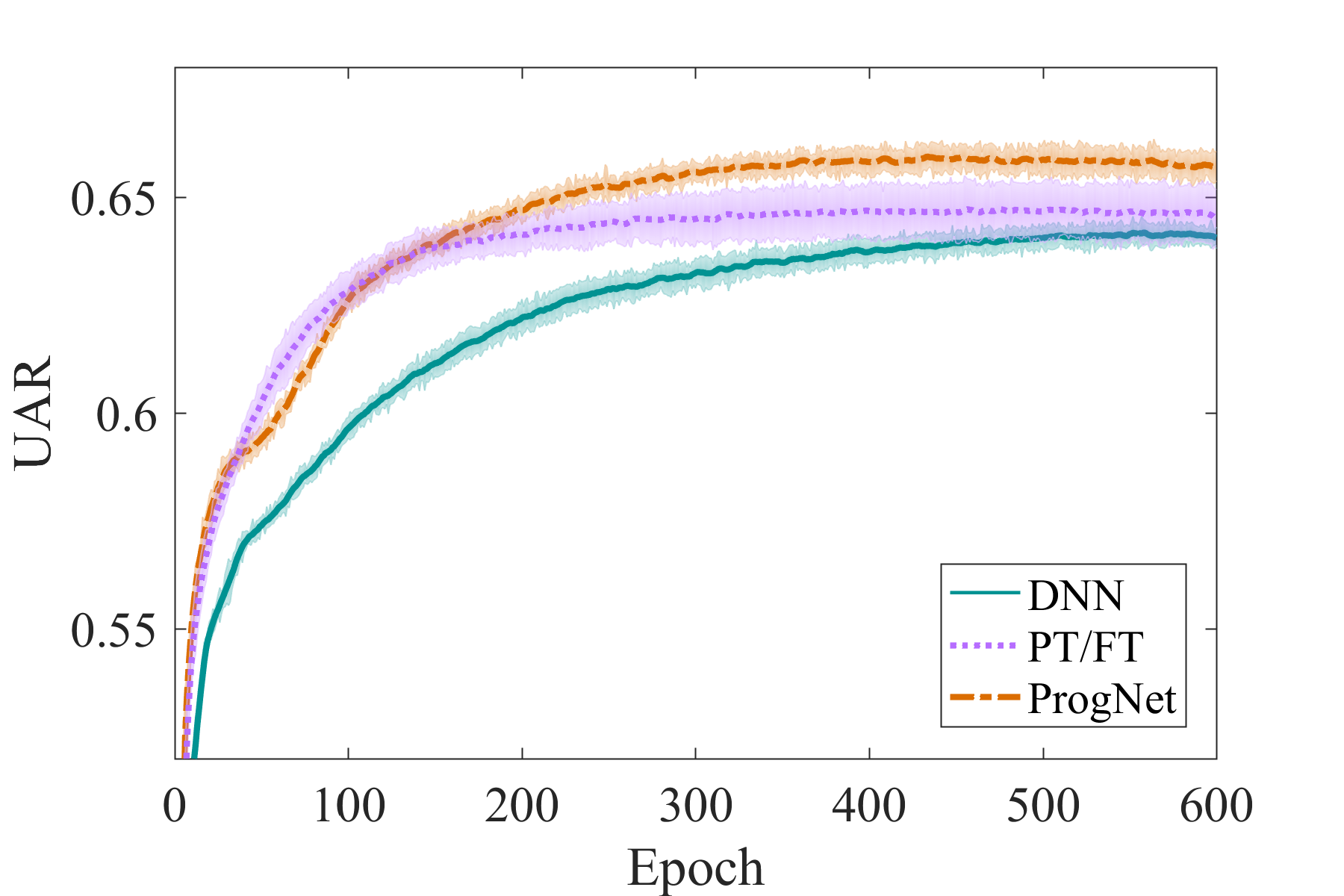}
\caption{\centering IEMOCAP}
\end{subfigure}
\begin{subfigure}[b]{0.46\linewidth}
\centering \includegraphics[width=\linewidth]{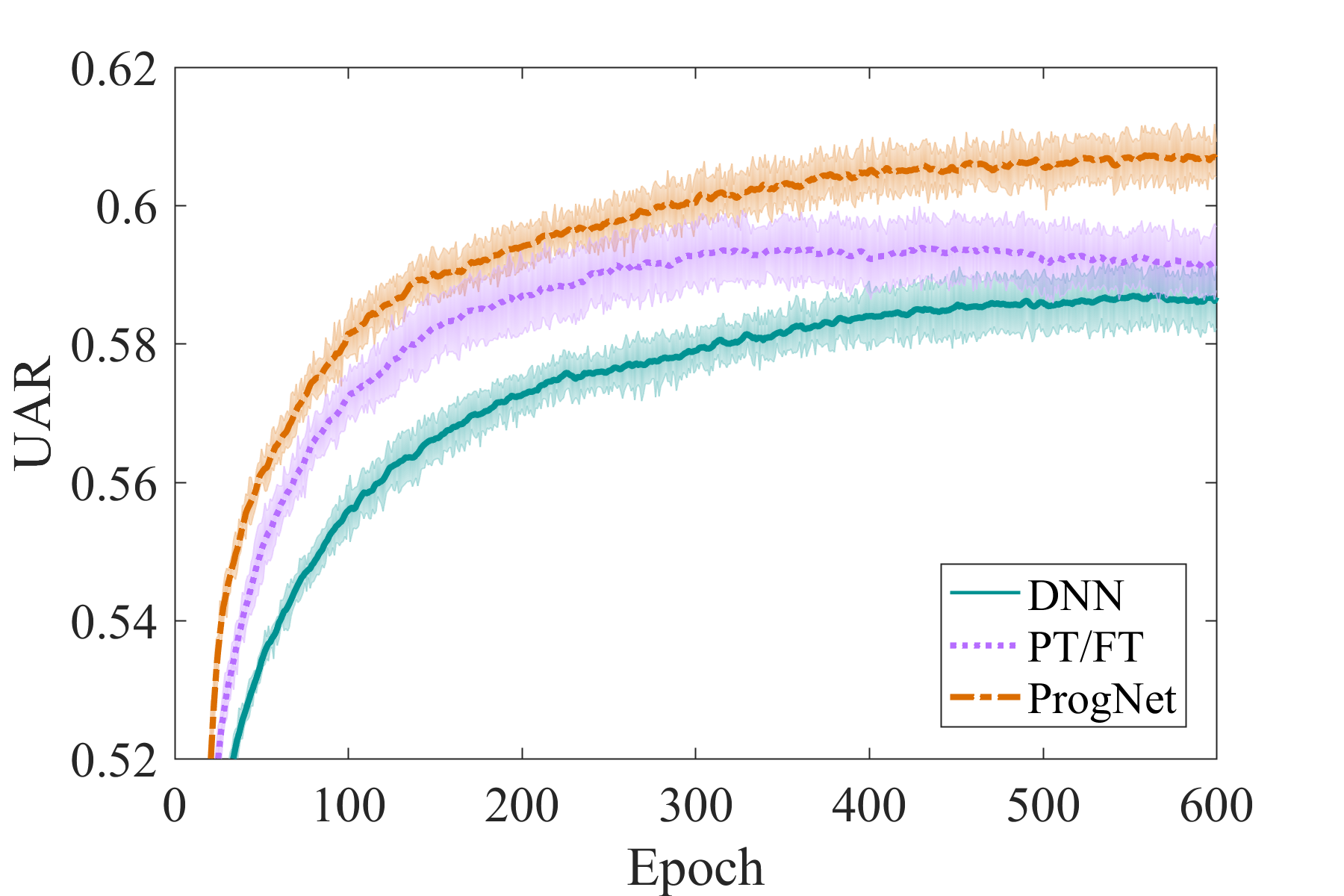}
\caption{\centering MSP-IMPROV}
\end{subfigure}
\caption{The learning curves of different methods when transferring representations from speaker to emotion. The regions around each curve show the standard deviation of the UARs found by averaging across the folds of each iteration.} \label{paraConvergence}
\end{figure*}
We report unweighted average recall (UAR) as our comparison measure. UAR is an 
unweighted accuracy that gives the same weights to different classes and is a 
popular metric for emotion recognition, used to account for unbalanced datasets~\cite{schuller2009interspeech}.
We evaluate the performance of the methods using a repeated ten-fold cross-validation scheme, as used in \cite{bouckaert2004evaluating}. 
The folds are stratified based on speaker ID. In each step of cross-validation, one
fold is used for testing, another is reserved for early stopping, and the remaining
eight folds are used for training. We repeat this evaluation scheme ten times, resulting in ten UARs for each iteration. We calculate the mean and standard deviation UAR within folds and report the mean of these statistics over all iterations. We perform significance tests using a repeated cross-validation paired $t$-test with ten degrees of freedom, as shown in \cite{bouckaert2004evaluating}, and note significance when $p < 0.05$.

\begin{table}[t]
\centering \caption{The hyper-parameters used in our experiments.}
\label{params_table}
\begin{tabular}{ c|c } 
Hyper-parameter & Value \\ 
\hline
number of layers & 4 \\ 
layers width & 256 \\ 
hidden activation function & sigmoid \\ 
output activation function & softmax \\ 
dropout rate & 0.5 \\ 
learning rate & 0.0005 \\ 
maximum number of epochs & 600 \\ 
\end{tabular}
\vspace{-5mm}
\end{table}


\section{Paralinguistic Experiments}
\subsection{Experimental Setup}
In the first set of experiments, we investigate the effectiveness of transferring knowledge from speaker or gender recognition to emotion recognition using the three methods mentioned above. In this section, we first report UAR of the systems on both IEMOCAP and MSP-IMPROV. We analyze the learning curves to compare the convergence behaviors of the systems. Prior work demonstrated that using the weights of a pre-trained model to initialize a new model to be trained on a related task can increase convergence speed
\cite{rusu2016progressive}.

\subsection{Results}
\begin{table}
\caption{Paralinguistic experimental results comparing different techniques 
for transferring knowledge from speaker/gender to emotion. Mean and standard 
deviation UARs are given for each method. A cross shows a result is 
significantly better than the other two methods for a given task, while an asterisk notes results significantly better than a standard DNN. The mean within-fold standard deviations are shown.}
\begin{tabular}{ c|c|c|c } 
Source & Method & IEMOCAP & MSP-IMPROV \\ 
\hline
N/A & DNN         & 0.640$\pm$0.017 & 0.584$\pm$0.022 \\ 
Speaker & PT/FT   & 0.645$\pm$0.017 & 0.592$\pm$0.018 \\ 
Speaker & ProgNet & \bf{0.657$\pm$0.018$^{*\dagger}$} &                                     \bf{0.605$\pm$0.021$^{*\dagger}$} \\ 
Gender & PT/FT    & 0.640$\pm$0.016 & 0.586$\pm$0.021 \\ 
Gender & ProgNet  & 0.642$\pm$0.018 & \bf{0.593$\pm$0.022$^*$} \\ 
\end{tabular}
\label{paraResultsTable}
\centering
\end{table}

\begin{figure*}[t]
\centering
\begin{subfigure}[b]{0.46\linewidth}
\centering \includegraphics[width=\linewidth]{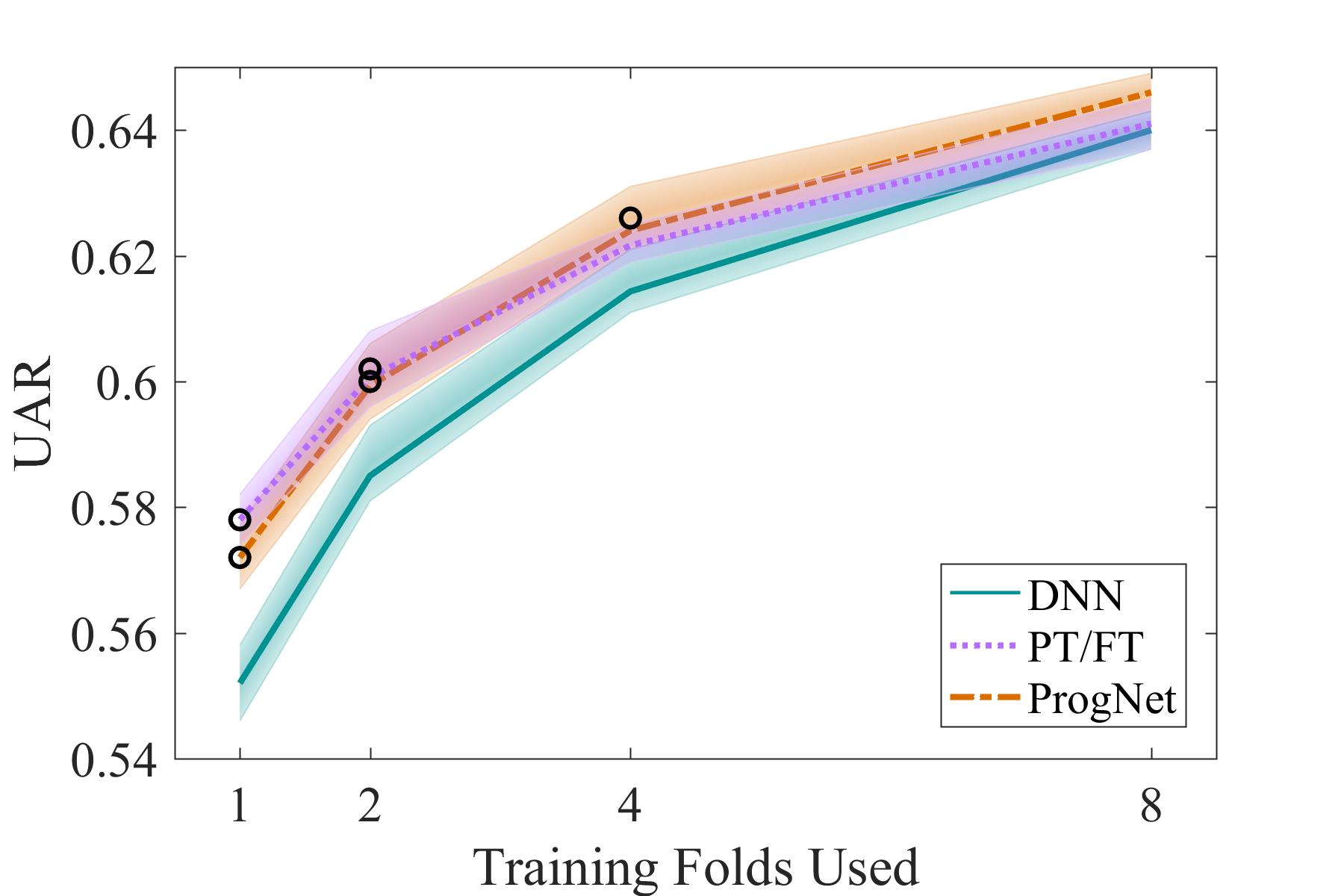}
\caption{\centering Source: MSP-IMPROV; Target: IEMOCAP}
\end{subfigure}
\begin{subfigure}[b]{0.46\linewidth}
\centering \includegraphics[width=\linewidth]{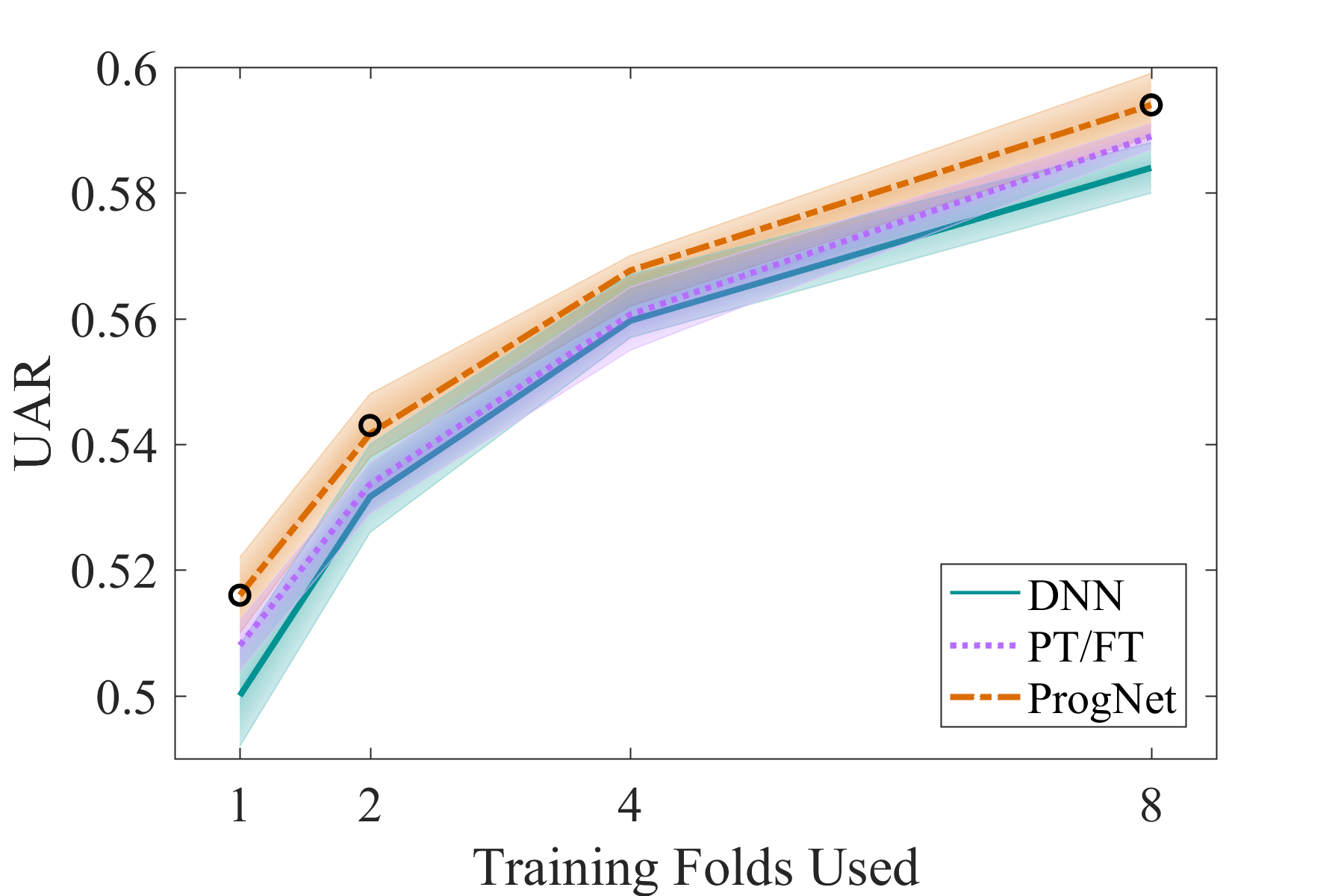}
\caption{\centering Source: IEMOCAP; Target: MSP-IMPROV}
\end{subfigure}

\caption{Cross-dataset experimental results under different amounts of training folds used (out of 8 total available training folds). Each test is run for ten iterations with different random folds to control for variations in selected data. All experiments use emotion as the source and target label. The regions around each curve show the standard deviation of the UARs found by averaging across the folds of each iteration. Circles mark results that are statistically significantly better than DNN.}
\label{crossDatasetResults}
\end{figure*}

Table \ref{paraResultsTable} summarizes the results obtained from speaker-emotion and gender-emotion transfer learning. When transferring from speaker recognition to emotion recognition, ProgNets significantly outperform both standard DNN ($p = 2.6$E-3) and PT/FT ($p = 2.0$E-2) for IEMOCAP and both standard DNN ($p = 8.8$E-4) and PT/FT ($p = 1.2$E-2) for MSP-IMPROV. The PT/FT system slightly outperforms the standard DNN, but the improvement is not significant. This suggests that ProgNets can efficiently incorporate representations learned by speaker recognition systems into emotion recognition ones, but PT/FT cannot leverage this knowledge as effectively.

The performance of transferring gender information using ProgNets is not consistent between IEMOCAP and MSP-IMPROV.
ProgNets significantly ($p =$ 1.2E-2) outperform the standard DNN when transferring knowledge from gender recognition to emotion recognition in the case of MSP-IMPROV, but not IEMOCAP. We hypothesize that this is due to the stronger gender recognition performance on MSP-IMPROV. Gender recognition UAR on MSP-IMPROV and IEMOCAP are $98.1\%$ and $93.1\%$, respectively. For both datasets, PT/FT is not effective at transferring gender information and performs no better than the standard DNN.

Figure \ref{paraConvergence} shows learning curves of the three reported systems for the specific case of transferring speaker knowledge to an emotion detection system. The figure shows that the PT/FT system reaches its best solution faster than the other
two methods. The PT/FT system, however, achieves lower final performance than that of ProgNets.
The learning curves of both transfer learning systems (i.e., PT/FT and ProgNet) start with a larger slope compared to the standard DNN. This slope vanishes quickly in PT/FT (after approximately $150$ epochs), but the slope for the ProgNet preserves a positive value up to approximately $400$ epochs. We hypothesize that this vanishing slope is due to PT/FT's inability to effectively incorporate representations learned for solving the source task.


\section{Cross-Dataset Experiments}
\subsection{Experimental Setup}
In this set of experiments, we explore transfer learning as a way to improve emotion recognition using an existing emotion model. 
In this experiment, the model is trained on the source dataset and is then adapted (PT/FT and ProgNet) to the target dataset.  The standard DNN is trained only on the target dataset.
We examine the impact of transfer learning when the target training data size is small by using different subsets of the training folds: 8, 4, 2, and 1. Previous work has shown that transferring knowledge from a large source data set to a smaller target datasets can be beneficial~\cite{das2015cross}. The source model is always trained using the full source dataset (all eight folds). We perform transfer learning by first treating IEMOCAP as the source and MSP-IMPROV as the target and we reverse the source/target designations. 

\subsection{Results}

Figure \ref{crossDatasetResults} shows a summary of the results when transferring across different corpora. ProgNet significantly outperforms the standard DNN when transferring from MSP-IMPROV to IEMOCAP for training fold sizes of 1 ($p = 2.0$E-3), 2 ($p = 1.1$E-2), and 4 ($p = 1.6$E-2) and when transferring from IEMOCAP to MSP-IMPROV for training fold sizes of 1 ($p = 5.0$E-3), 2 ($p = 3.2$E-2), and 8 ($p = 3.6$E-2). PT/FT only achieves significant improvement versus the standard DNN baseline when transferring from MSP-IMPROV to IEMOCAP for training fold sizes of 1 ($p =7.1$E-4) and 2 ($p =1.0$E-2).

Because ProgNet has a larger number of weights to transfer knowledge, it is most beneficial when the target dataset is larger, compared with PT/FT. We hypothesize that this is what causes PT/FT to perform better in cases of smaller training fold amounts (1 and 2) on a smaller target dataset (IEMOCAP). This indicates that in some cases of small data, PT/FT may still be the better choice. However, in cases where the size of the target dataset is sufficient, ProgNet can effectively utilize the previous task representation better than PT/FT.


\section{Conclusion}
 Transfer learning provides a method for using additional paralinguistic data, such as speaker ID, as well as a technique for combining models trained on different datasets.
This paper demonstrates the usefulness of progressive neural networks for this task. While pre-training a DNN on a source dataset has been previously used for transferring knowledge between tasks, progressive neural networks provide an alternative
way of avoiding the forgetting effect by allowing the network to retain representations learned for solving the original task.
ProgNets significantly outperformed the standard DNN and PT/FT networks when transferring knowledge between speaker identity and emotion.  This suggests the utility of ProgNets. We also demonstrated that ProgNets can provide significant improvements for gender to emotion transfer tasks and dataset transfer tasks when compared to systems that do not utilize source information.

In this work, we concentrated on transferring knowledge across paralinguistic tasks and datasets. However, 
future work will explore the utility of 
transferring knowledge across both simultaneously, even when the datasets have different characteristics (e.g. telephone vs. high fidelity speech). Speech and speaker recognition datasets tend to contain more training utterances than emotion datasets and contain paralinguistic information, such as subject ID. This data could be used to train a classifier for different types of speakers (e.g., accent, demographics). This model could then be transferred to emotion to account for person-specific variations tied to phenomena outside of emotion.
This would allow for the transfer of knowledge from multiple datasets or paralinguistic tasks. 
Future work will explore these ideas in order to better augment emotion classification with existing data.

\section{Acknowledgement}
This work was partially supported by IBM under the Sapphire project. We would like to thank Dr. David Nahamoo and Dr. Lazaros Polymenakos, IBM Research, Yorktown Heights, for their support.
  




\bibliographystyle{IEEEtran}

\bibliography{mybib}

\end{document}